\documentclass{article}
\usepackage[margin=1.25in]{geometry}
\usepackage[utf8]{inputenc}
\usepackage{caption}
\usepackage{graphicx}
\usepackage{subfig}
\usepackage{amsmath}
\usepackage{hyperref}

\title{Contextual Recurrent Neural Networks}

\author{Sam Wenke, Jim Fleming\\ \small{\{sam, jim\}@fomoro.com}}
\date{January 2019}

\begin{document}
\maketitle

\begin{abstract}
There is an implicit assumption that by unfolding recurrent neural networks (RNN) in finite time, the misspecification of choosing a zero value for the initial hidden state is mitigated by later time steps. This assumption has been shown to work in practice and alternative initialization may be suggested but often overlooked. In this paper, we propose a method of parameterizing the initial hidden state of an RNN. The resulting architecture, referred to as a \textit{Contextual RNN}, can be trained end-to-end. The performance on an associative retrieval task is found to improve by conditioning the RNN initial hidden state on \textit{contextual information} from the input sequence. Furthermore, we propose a novel method of conditionally generating sequences using the hidden state parameterization of \textit{Contextual RNN}.
\end{abstract}

\section{Introduction}

The initialization method of the hidden state of recurrent neural networks (RNN) is typically defaulted to a constant value, most commonly equal to zero. This is due to RNNs having the characteristic of accumulating information over time, producing outputs as a function of the hidden state and input at each time step. Interestingly, by choosing an arbitrary constant value for the initialization of the hidden state the RNN will learn to correct for discrepancies in the output credited to the initial value. In this paper, we show that the rate of the initial hidden state correction can be improved by learning a value conditioned on \textit{contextual information} related to the task.

Regardless, RNNs have been successfully applied to many hard generalization tasks such as natural language processing \cite{Yang2017BreakingTS} and reinforcement learning \cite{Espeholt2018IMPALASD}\cite{Ba2016UsingFW}. Many successful works that use RNNs rely on the underlying dynamics, capacity, and generalization of the hidden state, such as the use of long short-term memory (LSTM) \cite{Hochreiter1997LongSM} and fast weights (FW) \cite{Ba2016UsingFW} units. While many successes have been achieved with constant zero initialization of the hidden state, further gains may be found by learning and conditioning the initialization of the hidden state.\footnote{Code for this paper is publicly available at \url{https://github.com/fomorians/contextual_rnn}}

\subsection{Preliminaries}

\subsubsection{Recurrent Neural Networks}
A recurrent neural network (RNN) is a neural network that consists of a hidden state $h$ which operates on a variable length sequence $x = \{x_{1}, ..., x_{T}\}$. At each time step $t$, the hidden state $h_{t}$ of the RNN is updated by

\begin{equation}\label{eq:rnn}
    h_{t} = f(h_{t-1}, x_{t}), 
\end{equation}

\noindent{where $f$ is a non-linear activation function (such as a LSTM unit). The RNNs mentioned in this paper are optimized by stochastic gradient descent.}

The specification of the hidden state $h_{0}$ to a constant value is attributed to an RNNs ability to accumulate information over time. In other words, the RNN learns to eliminate the impact of a misspecification in the initialization on the outputs.

\subsubsection{Related Works}

One of the difficulties with finite unfolding in time is the proper initialization of the initial hidden state $h_{0}$. Fixing the hidden state to zero is a common choice; this has been shown to force the RNN to focus on the input sequence regardless of the initial choice of the hidden state, forcing the error in the initialization to diminish over time and training \cite{Zimmermann2005ModelingL}. Zimmermann et al. \cite{Zimmermann2005ModelingL} proposed a noise term (proportional to the back-propagated error) added to the initial hidden state to regularize and stabilize the dynamics of the RNN. As an alternative to an adaptive noise term, we propose that conditioning an initial hidden state distribution (using a reparameterization trick \cite{Kingma2013AutoEncodingVB}) on \textit{contextual information} related to the task can provide a better foundation for end-to-end handling of the underlying instabilities.

This \textit{contextual information} can be (but is not limited to) an encoded summary of an input sequence \cite{ChoM2014LearningPhrase, Bahdanau2014NeuralMachine}, a coarse image with multiple objects \cite{Ba2014MultipleOR}, or the first element of the input sequence. Ba et al. \cite{Ba2014MultipleOR} feed a down-sampled input image through a \textit{contextual network} to initialize the hidden state of an RNN and potentially provide sensible hints on where attention should be directed. Building on the ideas from this work and other similar sources, \textit{Contextual RNN} provide a general framework for learning the initial hidden state on a variety of forms of \textit{contextual information}.

\textit{RNN Encoder-Decoder} \cite{ChoM2014LearningPhrase} architectures use different initial hidden state initialization in the encoder and decoder RNN. Cho et al. \cite{ChoM2014LearningPhrase} and Bahdanau et al. \cite{Bahdanau2014NeuralMachine} initialize the hidden state of the encoder $h^{enc}_{0} = 0$ and the decoder $h^{dec}_{0} = tanh(W^{dec}_{h}h^{enc}_{T})$. By using the final hidden state of the encoder $h^{encode}_{T}$, the decoder is conditioned on \textit{contextual information} representing a summary of the input sequence. These works provided the foundations for the successful \texttt{seq2seq} framework using RNN. In the spirit of the initialization of the decoder RNN, \textit{Contextual RNN} proposes a potential alternative initialization for the encoder RNN as opposed to using a zero value.

More recent work uses attention mechanisms applied to the hidden state of RNN in various ways \cite{Ba2016UsingFW, Bahdanau2014NeuralMachine}. These methods attempt to increase both the long- and short-term memory capacities and handle very different time scales and alignments of sequential data. Bahdanau et al. \cite{Bahdanau2014NeuralMachine} use a non-linear \textit{context} function $q$ applied to the hidden states over time ($q(\{h^{dec}_{1}, h^{dec}_{2},...h^{dec}_{T}\})$) to model longer sequences and jointly learn alignments for language translation. Contrary to the decoder context used by this work, the FW formulation from Ba et al. \cite{Ba2016UsingFW} forces the RNN to iteratively attend to the recent past by a scalar product between the current hidden state $h_{t}$ and the previous hidden states $h_{t-1}$. In light of the successes of the attention mechanism applied to the hidden states formulated by the FW RNN, \textit{Contextual RNN} can be used to initialize the hidden state and improve upon the convergence speed on the associative retrieval task introduced by \cite{Ba2016UsingFW}.

\subsection{Our Contributions}

In this paper, we propose a method of parameterizing the initial hidden state of an RNN. The resulting architecture, referred to as a \textit{Contextual RNN}, can be trained end-to-end. A neural network called the \textit{context network} produces the initial hidden state of an RNN and is capable of improving the performance on multiple tasks. Initially, we qualitatively demonstrate that \textit{Contextual RNN} improves convergence speed on an associative retrieval task compared to the vanilla FW. Next, we describe a novel method to generate sequences by sampling hidden states conditioned on a regression quantity. Finally, we compare and summarize the results and generated samples trained on 1-D linear cosine decay. 

\section{Methods}

We explore the use of \textit{contextual information} taken from the input sequence to improve on the generalization of missing sequential key-value pairs. Furthermore, we propose a novel method of generating sequential data given \textit{contextual information} in the form of classification labels or regression quantities. 

\subsection{Contextual Recurrent Neural Networks}
We formally define a \textit{Contextual RNN} as an RNN with a learned initial hidden state conditioned on \textit{contextual information}. Using Eq. \ref{eq:rnn}, we outline various ways to initialize the initial hidden state,

\begin{subequations}
    \begin{align}              \label{eq:constant_state}
        h_{0} & = constant, \\ \label{eq:variable_state}
        h_{0} & = variable, \\ \label{eq:contextual_state}
        h_{0} & = g(context),
    \end{align}
\end{subequations}

\noindent{where $h_{0}$ can hold a $constant$ value (most commonly $h_{0} = 0$), a $variable$, or a neural network $g$.}

When the initialization (\ref{eq:constant_state}) is chosen, the error of the specification is inherently ignored and the RNN is expected to overcome this error by accumulating information over time and training. Alternatively, we can choose the initialization (\ref{eq:variable_state}), a free parameter, to allow back-propagation to automatically correct for the error in the specification of the initial value over training. Although, this method does not scale well beyond online gradient descent (in this context, this means the batch size is 1). Therefore, to promote the feasibility of this method with modern training approaches, the variable can be tiled (and noise can optionally be added to the copies of the variable) in order to be compatible with batches of data. Finally, the initialization (\ref{eq:contextual_state}) credits the specification error to \textit{contextual information} through a neural network $g$. When defining $g$ as a \textit{contextual neural network}, this method has several advantages and is the foundation of the \textit{Contextual RNN}.

\subsection{Experiments}

\subsubsection{Associative Retrieval Task}

\begin{figure}[h!]
    \centering
    \includegraphics[width=.65\textwidth]{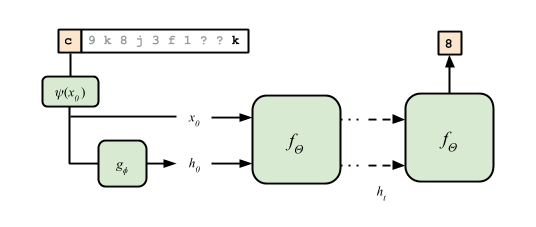}
    \captionof{figure}{The \textit{Contextual RNN} used in the associative retrieval task.}
    \label{fig:art_context_network}
\end{figure}

The associative retrieval task (ART) \cite{Ba2016UsingFW} tests the ability of an RNN to store and retrieve temporary memories. After generating a sequence of K character-digit pairs from the alphabet $\{a, b, c, ... z\}$ without replacement, one of the K different characters is selected at random as the query and the network must predict the target digit. In the input sequence, the query is presented directly after 2 $??$ characters. For example, the network is presented the input sequence \textbf{c9k8j3f1??k}, a query character for the sequence could be \textbf{k} and the corresponding target digit would be \textbf{8}.\footnote{The set of all other possibilities for this sequence are $\{\{c, 9\}, \{j, 3\}, \{f, 1\}\}$} A training set was used consisting of 100,000 samples, along with a validation set consisting of 20,000 samples.

We demonstrate the use of a single layer as the \textit{context network} $g$ with a shared character embedding $\psi{}(x_{0})$ that is conditioned on the first character in the input sequence (see Figure \ref{fig:art_context_network}). The models share the same architecture as in Ba et al. \cite{Ba2016UsingFW}, using a character embedding and a FW RNN with 50 hidden units. All the models were trained using batches of size 128 and the Adam optimizer \cite{Kingma2014AdamAM} with a learning rate of 0.001 for 200 epochs.

\begin{figure}[t]
    \centering
    \includegraphics[width=.475\textwidth]{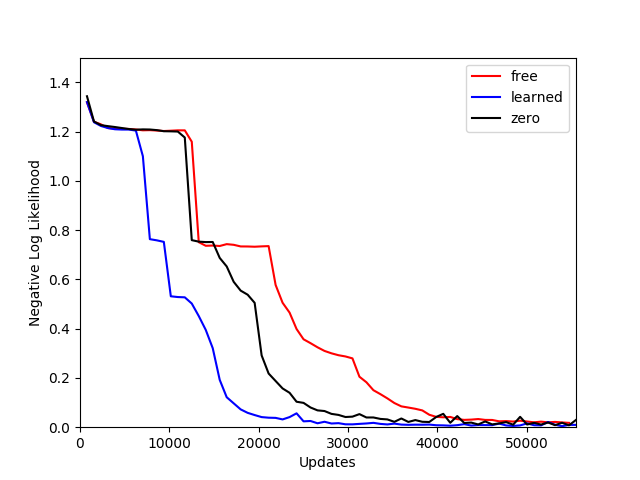}
    \includegraphics[width=.475\textwidth]{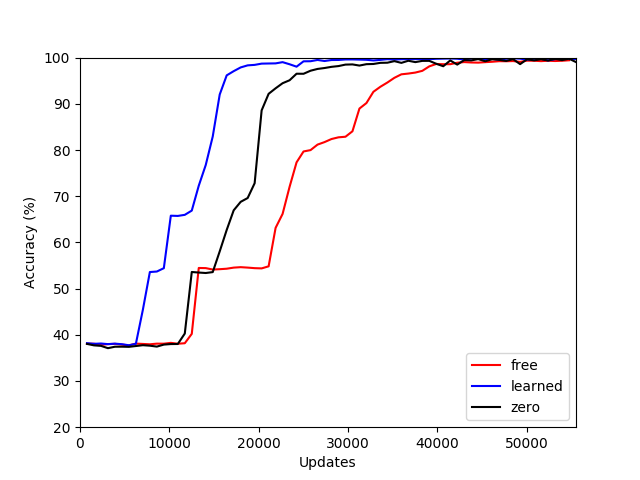}
    \captionof{figure}{Comparison of the validation log likelihood (\textit{left}) and accuracy (\textit{right}) on the associative retrieval task. The performance of the baseline \textit{zero} is improved by using a \textit{learned context state}.}
    \label{fig:art_results}
\end{figure}

From Figure \ref{fig:art_results}, we see that the \textit{Contextual RNN} form of the FW model (\textit{learned:} \ref{eq:contextual_state}) significantly outperforms the vanilla FW from Ba et al. \cite{Ba2016UsingFW} that use initialization methods (\textit{zero:} \ref{eq:constant_state}) and (\textit{free:} \ref{eq:variable_state}), without any fine-tuning.

\subsubsection{Linear Cosine Decay Task}

\begin{figure}[h!]
    \centering
    \includegraphics[width=.65\textwidth]{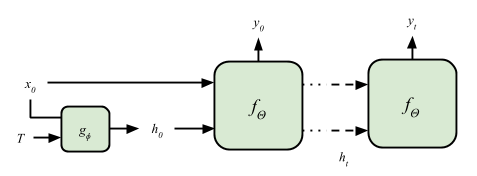}
    \captionof{figure}{The \textit{Contextual RNN} used in the linear cosine decay task.}
    \label{fig:lcd_context_network}
\end{figure}

\begin{figure}[t]
    \centering
    \begin{minipage}{.47\textwidth}
        \centering
        \includegraphics[width=.99\textwidth]{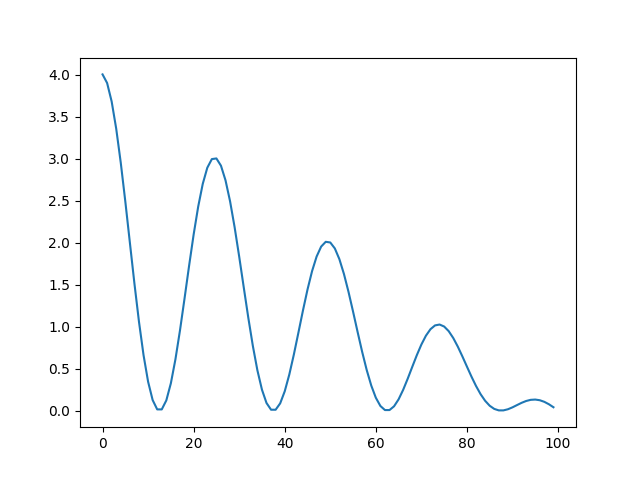}
        \captionof{figure}{Example of an LCD task where the constants are $\alpha{} = 0$ and $\beta{} = 0.001$, the period $T = 4.0$, the total decay time $t_{f} = 100$, and the initial value $x(0) = 4.0$.}
        \label{fig:lcd_example}
    \end{minipage}\hfill
    \begin{minipage}{.47\textwidth}
        \centering
        \includegraphics[width=.99\textwidth]{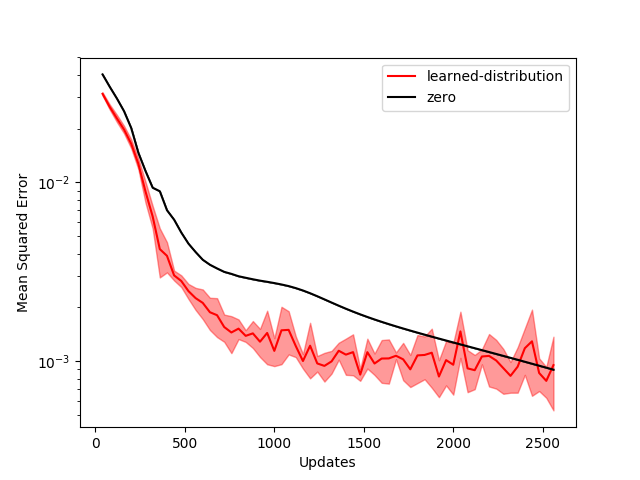}
        \captionof{figure}{Comparison of the validation squared error on
        the linear cosine decay task. The \textit{learned-distribution} model is trained and evaluated using 3 random seeds.}
        \label{fig:lcd_loss}
    \end{minipage}
\end{figure}

The linear cosine decay (LCD) task tests the ability of an RNN to interpolate and generalize to missing regression quantities. LCD\footnote{Our implementation of LCD uses the \texttt{tensorflow.train.linear\_cosine\_decay} \cite{tensorflow2015-whitepaper, LoshchilovH16a}.} is defined as

\begin{equation}
    x(t) = x(0)\Bigg[\bigg(\alpha + \frac{t_{f} - t}{t_{f}}\bigg) \bigg(\frac{1 + cos(\frac{2T\pi{}t}{t_{f}})}{2}\bigg) + \beta\Bigg],
\end{equation}

\noindent{where $t$ is the current time, $t_{f}$ is the total decay time, $T$ is the period, $alpha$ and $beta$ are constants, and $x(0)$ is the initial value. An example of a generated LCD sequence is shown in Figure \ref{fig:lcd_example}.}

In our experiments, the initial value is sampled from a uniform distribution $x(0) \sim \mathcal{U}(2,\,4)$, $\alpha{} = 0$ and $\beta{} = 0.001$. Training and validation sets are created using corresponding period sets, $T_{train}$ and $T_{validation}$. For each period $T$ in $T_{train} = \{0.5, 1.5, 2.5, 3.5, 4.5\}$, we sample 1000 LCD sequences $x$ of length $t_{f} = 25$ to create the the training set. We do the same for each period $T$ in $T_{validation} = \{1.0, 2.0, 3.0, 4.0\}$, sampling 500 LCD sequences $x$ of length $t_{f} = 25$ to create the validation set. The model is trained to produce the next input in the sequence by predicting the change in input $\delta{}x_{t} = x_{t} - x_{t-1}$ at each time step; $\delta{}x_{t}$ is added to the previous input to produce the next input $x_{t} = \delta{}x_{t} + x_{t-1}$. All values are scaled to be in the range $[-1, 1]$.

We demonstrate the use of the \textit{context network} $g$ with a hidden layer (with 50 units) that is conditioned on the initial value $x_{0}$ and period $T$ (see Figure \ref{fig:lcd_context_network}). The output of $g$ is used as the mean $\mu{}$ of a re-parameterized Normal distribution \cite{Kingma2013AutoEncodingVB} with a free variable for the scale $\sigma{}$, as in the equation

\begin{equation}\label{eq:rnn}
    h_{0} \sim{} \mathcal{N}(g(x_{0}, T),\,softplus(\sigma{}))
\end{equation}

\noindent{By conditioning a distribution on the initial value and period of the input sequence, we can implicitly generate a distribution of possible trajectories given the learned uncertainty under the initial hidden state given the data.}

\begin{figure}[t]
    \centering
    \includegraphics[width=.48\textwidth]{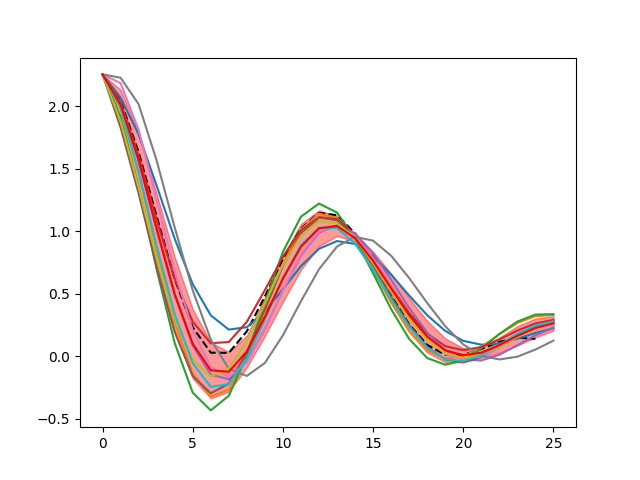}
    \includegraphics[width=.48\textwidth]{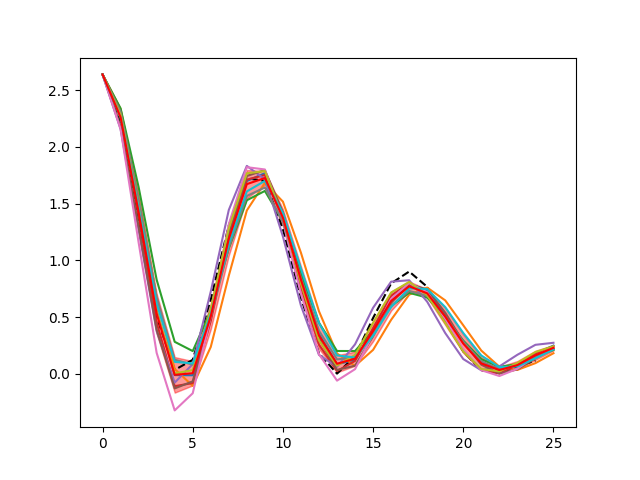}
    \captionof{figure}{Generated samples of periods $T = 2.0$ (\textit{left}) and $T = 3.0$ (\textit{right}) from the \textit{Contextual RNN} trained on the linear cosine decay task for 65 epochs.}
    \label{fig:lcd_generated}
\end{figure}

We chose to compare the \textit{zero} initial hidden state ($h_{0} = 0$) RNN and the \textit{Contexual RNN}. In order to properly compare the \textit{zero} initial hidden state with the \textit{Contextual RNN}, we append the period to each time step in the input sequences ($\{\{x_{0}, T\}, \{x_{1}, T\}, ... \{x_{t}, T\}\}$). Each model has a 128 unit LSTM followed by a final layer with a single unit corresponding to the output $\delta{}x_{t}$. All the models were trained using batches of size 128 and the Adam optimizer \cite{Kingma2014AdamAM} with a learning rate of 0.0002 for 65 epochs. The results on the validation set of during training are shown in Figure \ref{fig:lcd_loss}. After each epoch, we sampled 10 initial hidden states for a single example of each period in the validation set and show the results in Figure \ref{fig:lcd_generated}.

\section{Conclusion}

Conventional methods initialize the hidden state of RNN to zero regardless of the task. Other methods introduce \textit{contextual information} to both introduce new frameworks for solving variable length sequence problems \cite{ChoM2014LearningPhrase, Bahdanau2014NeuralMachine} and allowing RNN to summarize the input sequence \cite{Ba2014MultipleOR}. This paper provides a simple method of producing meaningful initial hidden states that are shown to improve performance on a range of tasks and introduces a novel generative model for sequences. These contributions explore the use of \textit{contextual information} to improve generalization of RNN.

\section{Acknowledgments}
We would like to thank Dan Saunders, Andrew Wang, Mike Qiu, Eddie Costantini and Danielle Swank at Fomoro for their constructive feedback on this report.

\bibliography{main}

\end{document}